\documentclass[lettersize,journal]{IEEEtran}
\usepackage{amsmath,amsfonts}
\usepackage{array}
\usepackage{textcomp}
\usepackage{stfloats}
\usepackage{url}
\usepackage{verbatim}
\usepackage{graphicx}
\usepackage{cite}

\usepackage{threeparttable}
\usepackage{makecell}
\usepackage{multirow} 
\usepackage{hhline}
\usepackage[table]{xcolor}
\usepackage{pifont}
\usepackage{subcaption}
\expandafter\def\csname ver@subfig.sty\endcsname{}
\usepackage[labelsep=period]{caption}
\usepackage{subfig}
\usepackage{colortbl}
\usepackage[ruled,linesnumbered]{algorithm2e}
\usepackage{tikz}
\usepackage{xcolor}
\usepackage{marvosym}
\definecolor{wine}{HTML}{CB0000} 
\definecolor{dark_green}{HTML}{009901}
\definecolor{code_Color}{HTML}{09A9EE}
\usepackage[pagebackref,breaklinks,colorlinks]{hyperref}
\newcommand{\circled}[1]{\tikz[baseline=(char.base)]{\node[shape=circle,draw,inner sep=0.2pt] (char) {#1};}}
\newcommand{\graycomment}[1]{\textcolor{black!50}{#1}}
\usepackage{todonotes}

\begin{document}

\title{Center-sensitive Kernel Optimization for \\ Low-cost Incremental Learning}

\author{Dingwen Zhang, Yan Li, De Cheng\textsuperscript{\Letter}, Nannan Wang, Junwei Han\textsuperscript{\Letter}, \textit{IEEE Fellow}}

\maketitle
\begin{abstract}
To facilitate the evolution of edge intelligence in ever-changing environments, this paper investigates low-cost incremental learning designed for devices with limited computational and memory resources. Existing low-cost training methods just focus on efficient training without considering the catastrophic forgetting, preventing the model from getting stronger when continually exploring the world. To address this challenge, we propose a simple yet effective low-cost incremental learning framework. Through an empirical analysis of the knowledge intensity of neural network kernel elements, we find the center kernel is critical for maximizing the knowledge intensity for learning new data, while freezing the other kernel elements would get a good balance between learning capacity and resource overhead. Building on this insight, we design a center-sensitive kernel optimization framework to significantly reduce the cost of gradient computation and back-propagation. Besides, a dynamic channel element selection strategy is introduced to facilitate a sparse orthogonal gradient projection, further reducing the optimization complexity by the knowledge explored from the new task data. Extensive experiments demonstrate the efficiency and effectiveness of our proposed method. Notably, our method achieves performance comparable to state-of-the-art incremental learning methods while requiring only $25\%$ of the computational cost and $8\%$ of the memory usage, highlighting its significant potential for incremental learning on resource-constrained devices.
\end{abstract}


\section{Introduction}
\label{Introduction}
The rapid advancements in embodied AI~\cite{wang2021wanderlust, khandelwal2022simple, kotar2022interactron} drive a growing demand for intelligent edge systems capable of adapting to ever-changing environments. This demand leads to the emergence of low-cost incremental learning, which requires edge devices to efficiently update their knowledge for a sequence of new tasks while preserving prior knowledge, all within very limited computational and memory resource budgets.

\begin{figure}[t]
	\centering	
        \includegraphics[width=\linewidth]{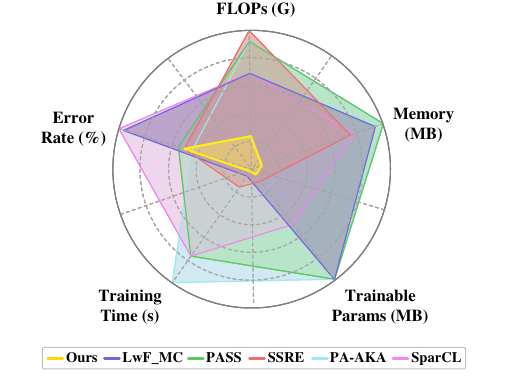}
	\caption{Our method demonstrates superior overall performance with \textit{lower computational cost (FLOPs)}, \textit{lower memory usage}, \textit{fewer trainable parameters}, \textit{shorter training time} and \textit{relatively higher classification accuracy (i.e., lower classification error rate)} compared with existing conventional and low-cost incremental learning methods.}
	\label{Intro}
\end{figure}

Recent progress in low-cost training introduces various methods to improve training efficiency in resource-constrained scenarios~\cite{alistarh2017qsgd, banner2018scalable, chen2020statistical, lin2022device, yuan2021mest}. These methods include techniques like generating sparse trainable parameter set~\cite{yuan2021mest} and applying gradient quantization to reduce back-propagation costs~\cite{alistarh2017qsgd, banner2018scalable, chen2020statistical, lin2022device}. However, when resource-constrained edge devices are deployed in ever-changing environments, they are further required to be able to learn efficiently from the changing data. Here, the challenge is that given the inherent resource limitation of edge devices, it is impractical to continuously accumulate new data and then retrain the entire model from scratch with all the old and new data. Another naive choice is to update the model only with the newly collected data, which, unfortunately, would lead to catastrophic forgetting and thus limit the usage of the model in future environments. In this context, it comes to the idea of bringing incremental learning schemes into the low-cost training framework to facilitate low-cost incremental learning. However, the fact is that although existing incremental learning techniques \cite{hou2019learning, hung2019compacting, mirzadeh2020understanding, bang2021rainbow, lin2022towards, tiwari2022gcr, kang2022forget, jin2022helpful, mohamed2023d3former} can alleviate catastrophic forgetting and achieve high performance for both the old tasks/data and the new ones, they typically rely on resource-intensive components such as historical model \cite{zhu2023self, shi2023prototype} or memory buffer \cite{bang2021rainbow, tiwari2022gcr}, making them unsuitable for resource-constrained devices. An initial attempt to address this is SparCL~\cite{wang2022sparcl}, which reduces resource demands by sparsifying trainable parameters. However, its unstructured parameter sparsification still requires all parameters to participate in gradient computations, limiting actual resource savings. 

To address the above challenges, we revisit the incremental learning scheme to explore whether it is possible to build a low-cost incremental learning framework that achieves strong learning performance while minimizing computational and memory costs. Our method focuses on more resource-friendly Convolutional Neural Networks (CNNs) architecture, and begins with an empirical study to investigate which parameters, specifically the elements within the convolution kernels, are pivotal in the learning process, thereby defining the “\emph{knowledge intensity}” of the kernel elements.

The empirical study is guided by two paradigms: sensitivity-induced assessment and amplitude-induced assessment. The sensitivity-induced assessment leverages data-related gradient information to evaluate the sensitivity of different parameters to incoming data, while the amplitude-induced assessment employs the inherent model weight to explore the contribution of each parameter to knowledge capturing. Based on the empirical study, we find that the central elements within the convolution kernel play a more pivotal role in learning knowledge from data, especially in the deeper layers of the network. This finding offers a valuable insight into developing efficient and effective incremental learning frameworks for resource-constrained edge devices. That is, selecting central kernel elements to maximize the plasticity of the network to learn on the new task data while freezing other kernel elements to keep stability on the old task knowledge would get a good balance between model incremental performance and training resource overhead.

Based on the above insight, we propose a novel technique called Center-sensitive Kernel Optimization (CsKO). As we know, directly optimizing the central kernel elements within the standard convolution kernel (e.g. $3 \times 3$ kernels) in the common operation manner would require gradient calculations across all parameters of the kernel and network layers, resulting in high computational and memory costs. In contrast, we decouple the central elements from the original convolution kernels into new $1 \times 1$  kernels, which are placed in an independent branch of the main network. The proposed CsKO mechanism enables the independent $1 \times 1$ kernel branch to undergo independent gradient computation and back-propagation. This approach significantly reduces the cost of gradient computation and back-propagation while preserving the essential properties needed for effective incremental learning.

To further alleviate the optimization burden with the upcoming new task data in an online manner, we introduce a Dynamic Channel Element Selection (DCES) strategy within the central kernel. As we know, the computational complexity of the orthogonal gradient projection\footnote{A technique widely used in the incremental learning framework to balance the stability and plasticity} \cite{wang2021training} is $O(N^3)$ to the number of channel elements due to the involved Singular Value Decomposition (SVD) of the covariance matrix. By selectively reducing the number of channel elements, the proposed DCES can mitigate the computational overhead of orthogonal gradient projection. Importantly, channel elements are selected based on their importance for learning new task data, ensuring that the reduction in learnable parameters does not hurt the plasticity of the model but further improves the stability. Consequently, DCES promotes the balance between model performance and training efficiency.

To sum up, the contributions of this paper are summarized as follows: 
\begin{itemize}
\item We revisit incremental learning mechanisms through an empirical study on the knowledge intensity of the kernel element. The study reveals that, in the convolution kernels of conventional convolutional neural network architectures, the central kernel elements play a significantly more pivotal role in the learning process compared to others.

\item Building on these findings, we propose a Center-sensitive Kernel Optimization (CsKO) mechanism that enables independent gradient calculation and back-propagation for newly formed $1\times 1$  central kernels. This is combined with a Dynamic Channel Element Selection (DCES) strategy to further enhance training efficiency. Together, these techniques offer a simple yet effective baseline for enabling low-cost incremental learning.

\item Extensive experiments on public benchmarks demonstrate that the proposed method achieves excellent incremental performance with significantly reduced computational cost, memory usage, trainable parameters and training time (as shown in Fig. \ref{Intro}), highlighting its strong potential for enabling incremental learning on resource-constrained edge devices.
\end{itemize}

The rest of this paper is organized as follows. In Section \ref{Related Work}, the related work about low-cost training and incremental learning is reviewed. In Section \ref{Empirical Study}, we conduct an empirical study on the knowledge intensity of kernel elements. We detail the proposed low-cost incremental learning framework in Section \ref{Methodology}. Section \ref{Experiments} presents extensive experimental results and discussions. Finally, we conclude the paper and provide some future research in Section \ref{Conclusion}.

\section{Related Work}~\label{Related Work}
\subsection{Low-cost Training} 
Low-cost training enables model optimization with low computational and memory overhead, making it feasible for resource-constrained devices to perform on-device model updates using local data to adapt to terminal environments. To meet the strict resource limitations of edge devices, some research is emerging to find ways to reduce the computational burden and memory footprint during training. 

The \textit{architecture modification} method proposed in \cite{cai2020tinytl} involves adding a small branch to the backbone model. During training, updates are confined to these small branches and the bias of the backbone model, with other parameters kept frozen. Some studies employ various \textit{sparse training} techniques \cite{parcollet2022zerofl, lin2022device} to dynamically prune the computation graph, thereby reducing the number of trainable parameters. However, a major limitation of most sparsity-based methods lies in their requirement to compute gradients for both pruned and non-pruned weights, which still consumes substantial memory resources. \textit{Gradient quantization} \cite{alistarh2017qsgd, banner2018scalable, chen2020statistical, lin2022device} is a promising technique for reducing training resource overhead. However, this technique faces challenges due to error accumulation from recursive gradient quantization across layers, which destabilizes the training process and leads to significant performance degradation. A recent work \cite{yang2023efficient} considers the key memory bottlenecks in training, i.e., activation and gradient, and proposes a novel gradient filtering method to generate an approximated gradient map with fewer unique elements. It significantly reduces the computational and memory overhead of backpropagation, facilitating efficient training on resource-constrained edge devices. 

Despite advancements in low-cost training, resource-constrained edge devices require the capability to continually learn new data while retaining old knowledge, strengthening the model to adapt to the ever-changing environments. In real-world scenarios, the common practice is to retrain the model from scratch once new data is available, which is infeasible for resource-constrained edge devices. On the other hand, simply updating or fine-tuning models with only the new data can lead to catastrophic forgetting, where the model loses previously acquired knowledge.

\subsection{Incremental Learning}
Incremental learning is the process of continually learning a series of new tasks while retaining knowledge from previously learned tasks, with the primary goal being the mitigation of catastrophic forgetting. Past efforts to alleviate catastrophic forgetting can be broadly classified into three categories: regularization-based, replay-based and structure-based methods.

\textit{Regularization-based methods} \cite{kirkpatrick2017overcoming, zenke2017continual, mirzadeh2020understanding, lin2022towards} add penalty terms to the objective function or use techniques like knowledge distillation to constrain the impact of parameter updates on old tasks. Based on the objective of regularization, these methods can be categorized into two types: weight regularization and function regularization. Weight regularization \cite{benzing2022unifying, liu2018rotate, park2019continual, lin2022towards} selectively penalizes changes in network parameters by incorporating a penalty term into the loss function, which penalizes changes according to each parameter's importance to old tasks. Function regularization \cite{li2017learning, iscen2020memory, dhar2019learning, wang2022continual} targets the intermediate or final outputs of the model, typically using the previously learned model as the teacher and the current model as the student, simultaneously implementing knowledge distillation (KD) \cite{gou2021knowledge} to mitigate catastrophic forgetting. 

\textit{Replay-based methods} \cite{lopez2017gradient, hou2019learning, bang2021rainbow, tiwari2022gcr} store a sample or prototype memory bank from previous tasks and jointly train the model with new tasks to maintain old knowledge. Early efforts typically store several old training samples in a small memory buffer. Since samples are raw images, directly saving a set of instances can consume significant memory costs. To address this, some approaches are proposed to construct memory-efficient replays. For instance, \cite{iscen2020memory} suggests storing extracted features of samples to alleviate the memory burden. Similarly, \cite{zhao2021memory} proposes maintaining low-fidelity images instead of raw images. 

\textit{Structure-based methods} \cite{mallya2018piggyback, mallya2018packnet, hung2019compacting, kang2022forget, jin2022helpful} reduce interference between tasks by isolating existing model parameters or allocating additional parameters for new tasks. Parameter isolation \cite{mallya2018packnet, kang2022forget, jin2022helpful} features a static network architecture to continuously learn tasks, allocating disjoint subsets of parameters for each task to prevent catastrophic forgetting. However, due to the limited capacity of the network, the model tends to saturate as more incremental tasks are introduced. To alleviate this dilemma, some studies \cite{yan2021dynamically, RameshC22, YoonYLH18, hung2019compacting} explore dynamically expanding the network to better accommodate the learning of new tasks, such as adding a set of neurons or subnetworks. Nevertheless, the number of model parameters in these methods increases proportionally with the growing number of learned tasks. 

These methods improve incremental learning performance through various computational and memory-intensive strategies, such as saving old samples or models, extending networks, etc., which makes them impractical to train on resource-constrained devices.

\begin{figure*}
	\centering
	\begin{subfigure}{0.47\linewidth}
		\includegraphics[width=\linewidth]{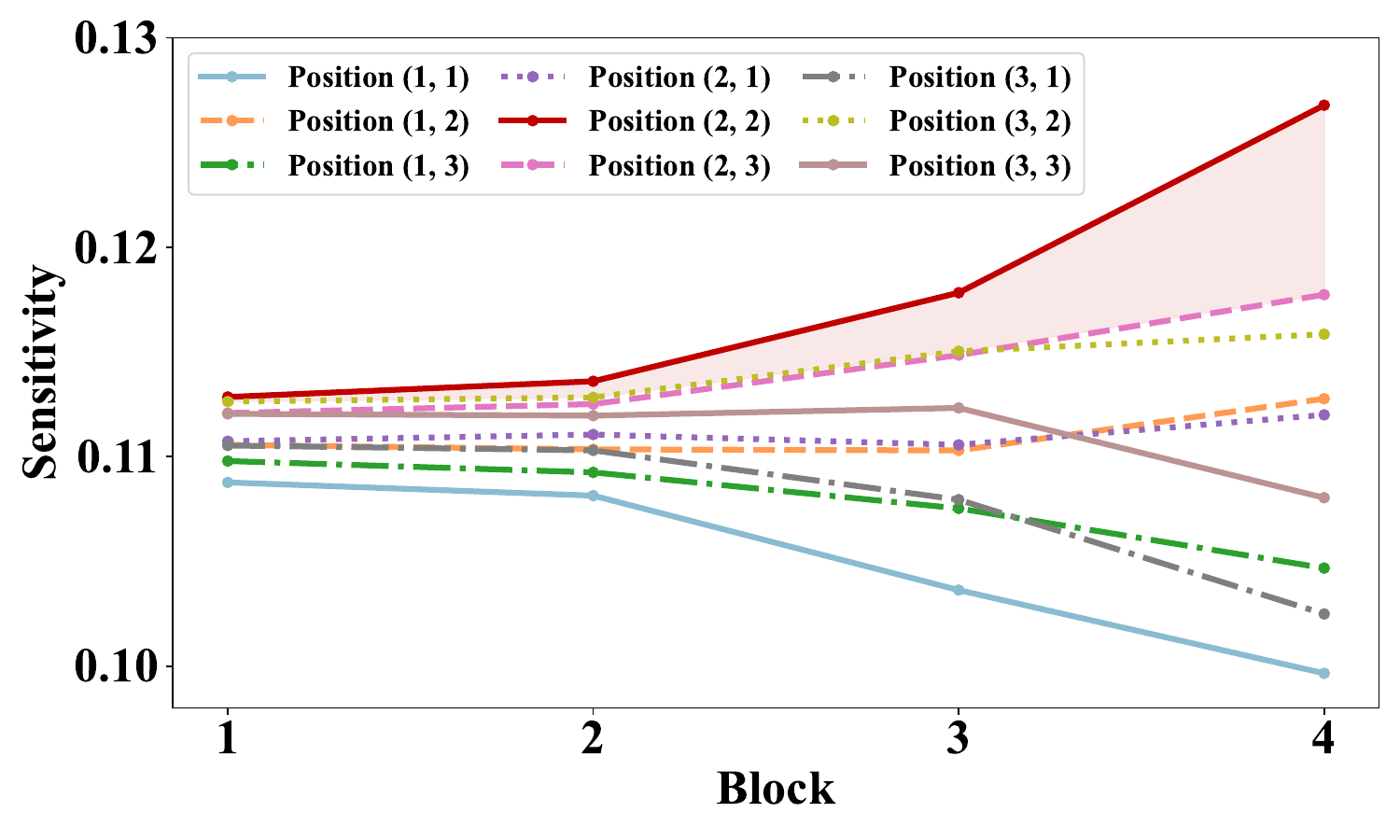}
	\end{subfigure}
	\begin{subfigure}{0.47\linewidth}
		\includegraphics[width=\linewidth]{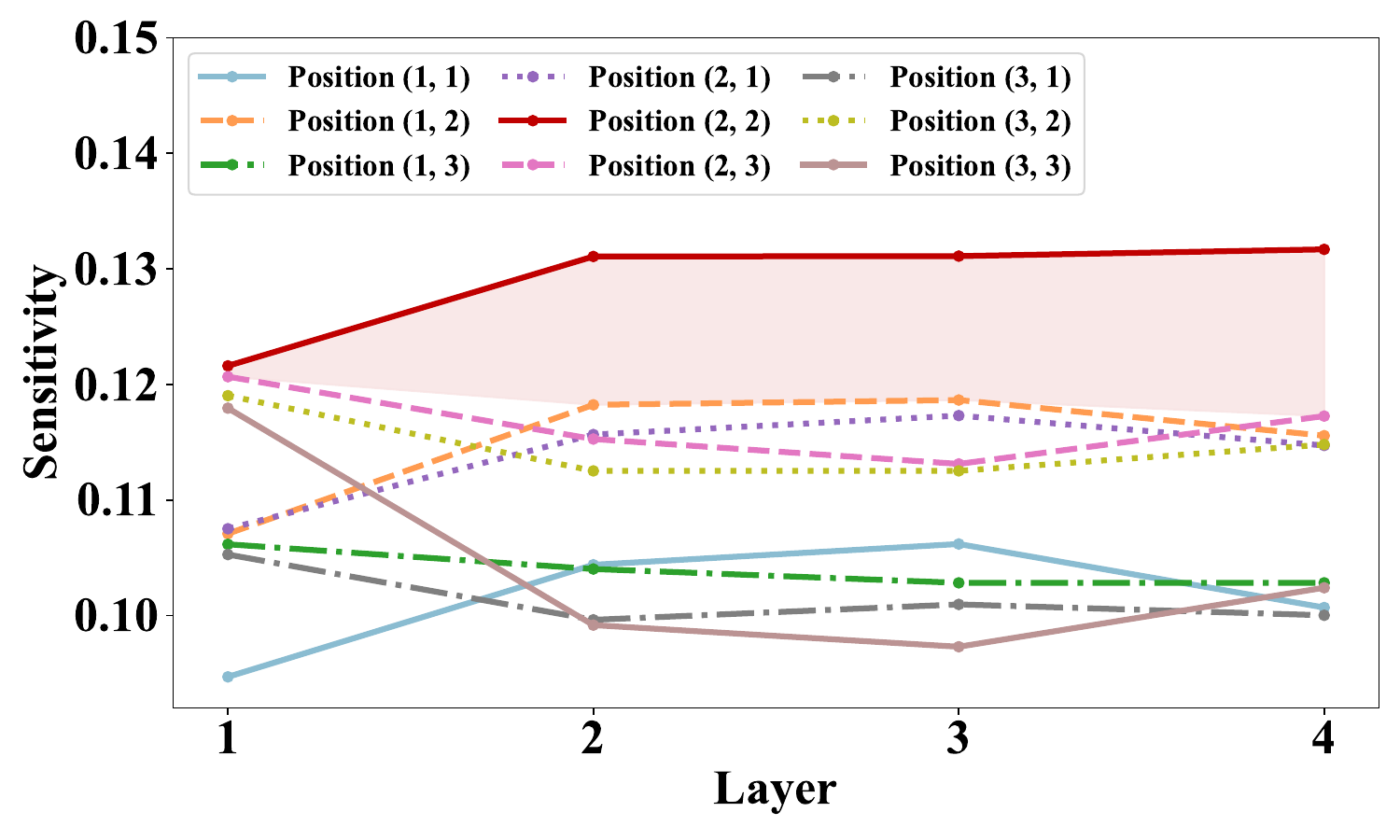}
	\end{subfigure}
	\begin{subfigure}{0.47\linewidth}
		\includegraphics[width=\linewidth]{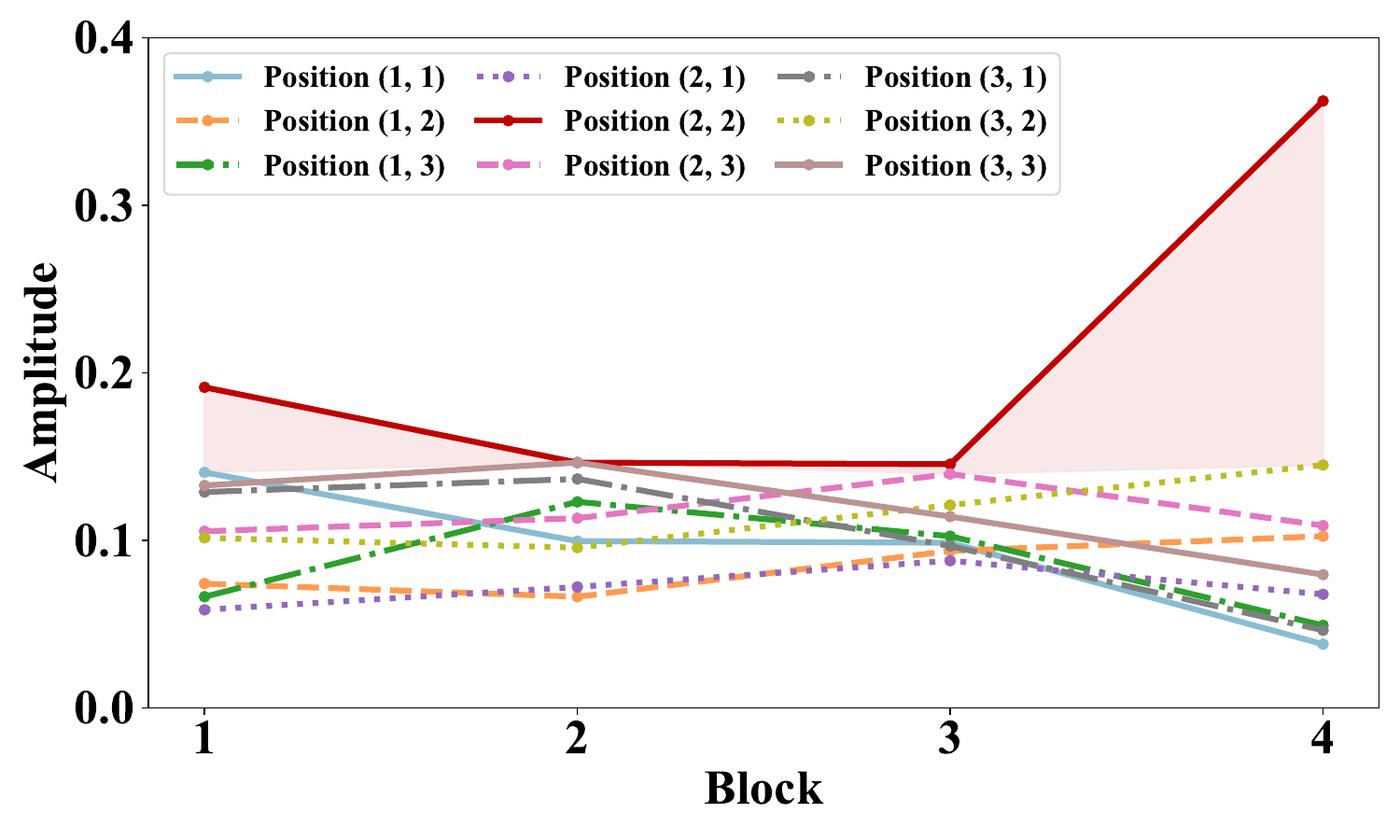}
	\end{subfigure}
	\begin{subfigure}{0.47\linewidth}
		\includegraphics[width=\linewidth]{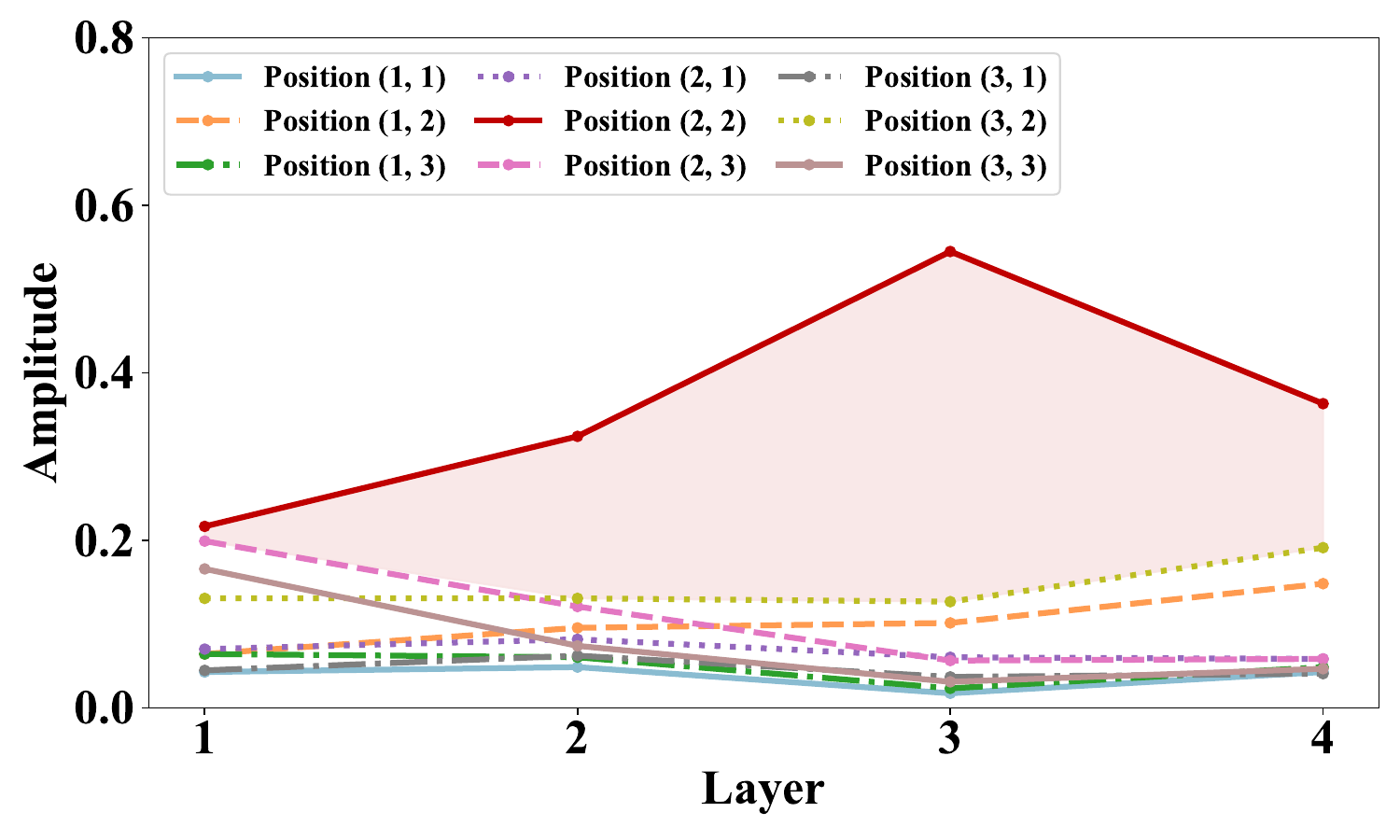}
	\end{subfigure}
	\caption{Results of sensitivity-induced (top) and amplitude-induced (bottom) knowledge intensity analysis using pre-trained ResNet-18 \cite{he2016deep} on the TinyImageNet \cite{le2015tiny} dataset. \textit{Left}: The sensitivity and amplitude of the $9$ positions in the $3 \times 3$ convolution kernel across all blocks within the network.  \textit{Right}: A detailed display of the sensitivity and amplitude across all layers in the last block.}
	\label{Pre_Analysis}
\end{figure*}

\section{Empirical Study on the Knowledge Intensity of Kernel Elements}~\label{Empirical Study}
Recent research has demonstrated that different parameters in pre-trained models exhibit varying contributions to downstream tasks \cite{he2023sensitivity, zheng2023regularized}. 
Some studies even suggest that more tunable parameters do not necessarily lead to better performance, and fine-tuning a subset of model parameters could usually achieve comparable or even better performances~\cite{ding2023parameter}. Motivated by this, we seek to evaluate which parameters, $i.e.$, the elements in the convolution kernels of a deep model, are more pivotal for learning new knowledge of upcoming training data. Specifically, we define this character as the \emph{knowledge intensity of kernel elements}.

\textbf{Sensitivity-induced knowledge intensity assessment.}
One way to measure the knowledge intensity of the kernel elements is to assess their importance for learning knowledge from the new task data. 
Define  $\mathbf{W}^{i} \in \mathbb{R}^{D\times C\times K\times K}$ as the convolution kernels of the $i$-th convolution layer in the pre-trained model, where $K\times K$ is the kernel size,  $C$ is the number of channels of the input feature map, and $D$ is the number of current filters ($i.e.,$ number of channels of output feature map by current convolution layer). We quantify the importance of specific kernel element $w^{i} \in \mathbf{W}^{i}$ by evaluating the effect of its changes on the loss $\mathcal{L}$, obtaining the sensitivity-induced knowledge intensity measurement. Given the data pair $(x, y)$ from the training data, we calculate the sensitivity score of $w^{i}$ by $S_{w^{i}} = |\mathcal{L}(x, y, \mathbf{W}^{i}) - \mathcal{L}(x, y, \mathbf{W}^{i}|w^{i}=\hat w^{i})|$, where $\hat w^{i}=w^{i}+\Delta w^{i}$, and the $\Delta w^{i}$ denotes the update for $w^{i}$. Then, following \cite{molchanov2019importance,he2023sensitivity}, the computation can be approximated by using the first-order Taylor series expansion: $S_{w^{i}} \approx |\frac{\partial\mathcal{L}}{\partial w^{i}}|$, where $\frac{\partial\mathcal{L}}{\partial w^{i}}$ represents the gradient on the kernel element $w^i$. 

To investigate the knowledge intensity of each kernel element, we accumulate the sensitivity scores of the kernel elements at the same location across both the channel and kernel dimensions as follows:
\begin{equation}
	S_{\mathbf{W}^{i}[u,v]} = \sum_{d=1}^{D}\sum_{c=1}^{C}|\frac{\partial\mathcal{L}}{\partial \mathbf{W}^{i}[d,c,u,v]}|,
\end{equation}
where $\frac{\partial\mathcal{L}}{\partial \mathbf{W}^{i}[d,c,u,v]}$ represents the kernel element gradient located at the position $(u, v)$ within the $c$-th channel of the $d$-th convolution kernel. For the conventional $3 \times 3$ convolution kernel, we analyze the sensitivity of $9$ spatial positions across all blocks, as shown in Fig. \ref{Pre_Analysis}. To better analyze the contribution of each position, we normalize the sensitivity scores of each kernel element: $S_{\mathbf{W}^{i}[u,v]}^{\text{*}}=\frac{S_{\mathbf{W}^{i}[u,v]}}{S_{\mathbf{W}^{i}}}$, where $S_{\mathbf{W}^{i}} = \sum_{u}\sum_{v}S_{\mathbf{W}^{i}[u,v]}$ represents the sum of sensitivity scores across all positions.

\textbf{Amplitude-induced knowledge intensity assessment.}
Another effective way to evaluate the knowledge intensity is the amplitude of the kernel element weight, where a larger amplitude of the weight indicates higher knowledge intensity of the corresponding kernel element after model training~\cite{chen2023run, ding2019acnet, shen2022prune}. Under this circumstance, we analyze the amplitude of the $9$ kernel elements within the $3 \times 3$ convolution kernels, as shown in Fig. \ref{Pre_Analysis}, where the amplitude-induced knowledge intensity of each kernel element is calculated as follows:
\begin{equation}
A_{\mathbf{W}^{i}[u,v]} = \sum_{d=1}^{D}\sum_{c=1}^{C}|\mathbf{W}^{i}[d,c,u,v]|,
\end{equation}
where $\mathbf{W}^{i}[d,c,u,v]$ denotes the weight at the $(u, v)$ position within the $c$-th channel of the $d$-th convolutional kernel.

As shown in Fig. \ref{Pre_Analysis}, the two assessment approaches reflect \textbf{a coincident finding}: \textit{The central kernel element, i.e. the one at position $(2, 2)$, always exhibits a higher knowledge intensity than others and its superiority tend to be overpowering in very deep network block and layers.} 

The above finding indicates that in the incremental learning task, if one can only select one kernel element to maximize the plasticity of the network to learn the new task data while freezing other kernel elements to keep stability on the old task knowledge due to the limitation on the computational and memory resource, the selected kernel element must be the central kernel element while the surrounding kernel elements are the frozen ones. Moreover, the finding even points out the best locations for facilitating incremental learning on the central kernel elements---the last three layers of the fourth block of the network---which is highly valuable for guiding the concrete design of the low-cost incremental learning framework.  

\section{Methodology}~\label{Methodology}
\label{Method}
\begin{figure*}[t]
	\centering
	\includegraphics[width=\textwidth]{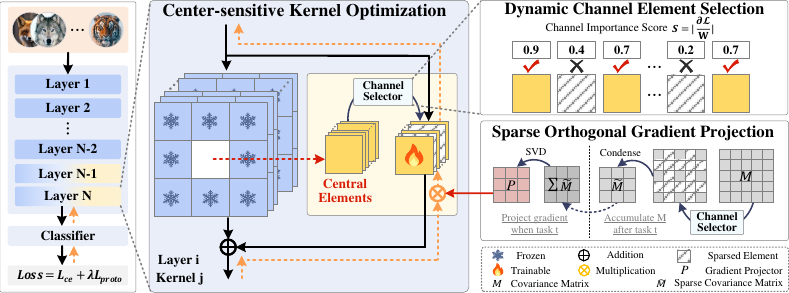}
	\caption{Illustration of the proposed method workflow. The central elements in the trainable layers are decoupled from the original convolution kernels into new $1 \times 1$ kernels placed on the side of the main network, which undergo independent gradient computation and back-propagation, significantly reducing training resource overhead. Then, the dynamic channel element selection strategy selects a proportion $s$ of central element channels that are more sensitive to the incoming new data, further alleviating the optimization burden. Besides, facilitated by the dynamic channel element selection, an efficient sparse orthogonal gradient projection is introduced to constrain parameter updates, effectively mitigating catastrophic forgetting.}
	\label{Framework}
\end{figure*}

In this section, we discuss the proposed low-cost incremental learning framework inspired by the above empirical analysis, which includes Center-sensitive Kernel Optimization (CsKO) mechanism and Dynamic Channel Element Selection (DCES) strategies. Fig. \ref{Framework} illustrates the workflow of the proposed method.

\subsection{Center-Sensitive Kernel Optimization}
The above finding provides a valuable insight for building the low-cost incremental learning framework, which involves selecting the central kernel elements to learn new data to maximize model plasticity, while keeping the remaining elements frozen to ensure model stability, achieving an optimal trade-off between model incremental performance and training resource consumption. However, conventional parameter optimization techniques calculate gradients across all parameters of the kernel, which results in the involvement of the non-central kernel elements during the back-propagation of the central kernel elements. Due to the inherent computation- and memory-intensive nature of backpropagation, this conventional optimization approach poses resource overhead challenges for incremental learning on resource-constrained devices. 

To address this challenge, we propose a Center-sensitive Kernel Optimization (CsKO) mechanism involving a \textbf{Center-sensitive Kernel Selection (CsKS)} collaborating with the \textbf{Center-decoupled Mechanism (CdM)} strategy. Instead of directly optimizing the central kernel elements within the convolutional kernel like conventional optimization methods, the proposed CsKO optimization mechanism decouples the central kernel elements from the original convolutional kernel to form a new $1 \times 1$ kernel, which is placed as an independent branch on the side of the main network. During the model optimization process, gradients are only allowed to flow through the independent $1 \times 1$ branch, while the remaining parameters are kept frozen, thereby substantially reducing the computational and memory overhead of backpropagation. 

Specifically, given the parameters of the $i$-th trainable convolution layer $\mathbf{W}^{i} \in \mathbb{R}^{D \times C \times K \times K}$, the CdM enables the central kernel elements be decoupled from $\mathbf{W}^{i}$, ultimately forming independent $1 \times 1$ kernel parameters $\mathbf{W}_{\alpha}^{i} \in \mathbb{R}^{D \times C \times 1 \times 1}$ and the remaining backbone parameters $\mathbf{W}_{\theta}^{i} \in \mathbb{R}^{D \times C \times K \times K}$:
\begin{equation}
\begin{aligned}
    \mathbf{W}_{\alpha}^{i}[d, c, 1, 1] &= \mathbf{W}^{i}[d, c, \lceil \frac{K}{2} \rceil, \lceil \frac{K}{2} \rceil], \\
    &\quad \forall d \in [1, D], c \in [1, C],
\end{aligned}
\label{eq:3}
\end{equation}
where $\lceil \cdot \rceil$ represents the round up of the element. The decoupled parameters $\mathbf{W}_{\alpha}^{i}$ are used for new knowledge learning, with the capability of computing gradients and performing backpropagation independently, while the remaining backbone parameters $\mathbf{W}_{\theta}^{i}$ are frozen to preserve the previously learned knowledge. The formalization of $\mathbf{W}_{\theta}^{i}$ is as follows: 
\begin{equation}
\begin{aligned}
    \mathbf{W}_{\theta}^{i}[d, c, u, v] = 
	\left\{
	\begin{array}{ll}
		0, &\mbox{if } u=v=\lceil \frac{K}{2} \rceil,\\	
		\mathbf{W}^{i}[d,c,u,v], &\mbox{otherwise},
	\end{array}
    \right. \\
    \quad \forall d \in [1, D], c \in [1, C].
\end{aligned}
\label{eq:4}
\end{equation}

The proposed CsKO mechanism facilitates efficient element-level optimization in an effectiveness-equivalent manner, which offers an innovative perspective for low-cost training on resource-constrained edge devices.

\begin{figure*}{}
	\centering
	\includegraphics[width=\linewidth]{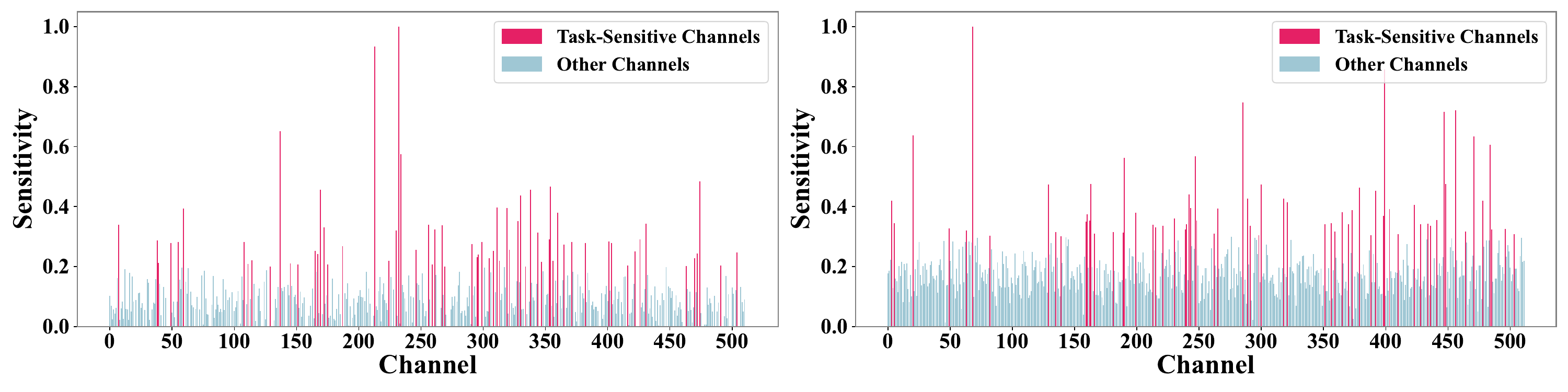}
	\caption{Visualization of the channel sensitivity analysis results for a trainable convolutional layer on CIFAR-100 dataset (left) and TinyImageNet (right) dataset. The result suggests that different channels of the convolutional layer exhibit varying levels of sensitivity to different data.}
	\label{Channel_Histogram}
\end{figure*}

\subsection{Dynamic Channel Element Selection} 
To further reduce the optimization burden with the upcoming new data in an online manner, we propose a \textbf{Dynamic Channel Element Selection (DCES)} strategy. This strategy aims to dynamically select channels based on their sensitivity to various incoming training data. Specifically, we analyze the contributions of individual channels to different input data (e.g., data from CIFAR-100 and TinyImageNet datasets) using a pre-trained ResNet-18 model. As shown in Fig. \ref{Channel_Histogram}, channels exhibit varying degrees of contribution to different task data, revealing distinct task-specific patterns. This observation motivates the design of the DCES strategy, which operates within the online learning process, facilitating more efficient and data-driven model optimization based on the incoming data stream. 

Specifically, for each given incremental learning task, an initial evaluation of the channel sensitivity to the task-specific data is conducted, which serves as the basis for dynamic channel element selection. Benefiting from the proposed center-sensitivity kernel optimization mechanism, the computation of channel sensitivity is confined to the decoupled $1 \times 1$ kernel parameters $\mathbf{W}_{\alpha}^{i} \in \mathbb{R}^{D \times C \times 1 \times 1}$.
The accumulated sensitivity scores for a given channel are computed as the combined sensitivity score:
\begin{equation}
	S_{\mathbf{W}_{\alpha}^{i}[c]} = \sum_{d=1}^{D}|\frac{\partial\mathcal{L}}{\partial \mathbf{W}_{\alpha}^{i}[d,c,1,1]}|.
\label{channel_score}
\end{equation}

Based on the channel sensitivity scores computed using Eq. \ref{channel_score}, the top $s \in [0, 1]$ proportion of channels with higher sensitivity to new data are selected for learning new knowledge, while the remaining channels are kept frozen to preserve previously learned knowledge. The proposed dynamic channel element selection mechanism further facilitates the balance between the model plasticity and training efficiency in low-cost incremental learning. 

Besides, the DCES strategy facilitates the development of a \textit{sparse orthogonal gradient projection}. The original “Orthogonal Gradient Projection” (OGP) effectively mitigates catastrophic forgetting without the need for additional parameter allocation or data storage, making it a more cost-effective choice for building incremental learning systems. However, the original OGP method incurs considerable computational and memory overhead primarily due to the Singular Value Decomposition (SVD) operation used to compute the gradient projection space. Specifically, for a matrix of size $N \times N$, the computational complexity of SVD is $O(N^3)$, and the memory complexity is $O(N^2)$. The proposed dynamic channel element selection mechanism inherently provides a potential solution to this challenge. Therefore, we develop an improved sparse orthogonal gradient projection strategy, which significantly alleviates the computational and memory overhead of the original OGP, making it more suitable for incremental learning on resource-constrained edge devices.

Specifically, the trainable $1 \times 1$ kernel parameters $\tilde{\mathbf{W}}_{\alpha}^{i} \in \mathbb{R}^{D \times (sC) \times 1 \times 1}$ after the DCES exhibit channel-wise sparsity with a sparsity factor $s \in [0, 1]$. Here, the input feature and output feature of the $i$-th layer are denoted as $\tilde{\mathbf{X}}^{i} \in \mathbb{R}^{B \times (sC) \times H \times W}$ and $\tilde{\mathbf{Y}}^{i}\in \mathbb{R}^{B \times D \times H' \times W'}$ respectively. The relationship between the input and output features is expressed as $\tilde{\mathbf{Y}}^{i} = \tilde{\mathbf{X}}^{i} \ast \tilde{\mathbf{W}}_{\alpha}^{i}$, where $\ast$ represents the convolution operation. This formulation indicates that only a subset of the channels in the input features are actively involved in the convolution process. To compute the gradient projection space, the covariance matrix is derived from the input features $\tilde{\mathbf{X}}^{i} \in \mathbb{R}^{B \times (sC) \times H \times W}$, i.e., $\tilde{\mathbf{M}} = \text{Covar}(\tilde{\mathbf{X}}^{i}, \tilde{{\mathbf{X}}^{i}}^{\top})$, where $\tilde{\mathbf{M}} \in \mathbb{R}^{(sC) \times (sC)}$, \text{Covar}$(\cdot)$ represents the covariance operation, and $\top$ denotes transposition. Notably, the covariance matrix is sparse, which enables performing SVD on the sparse covariance matrix, thereby facilitating more efficient computation of the gradient projection space. 

Theoretically, in conventional orthogonal gradient projection-based methods, the memory complexity of performing SVD on a dense covariance matrix $\mathbf{M} \in \mathbb{R}^{C \times C}$ is $O(C^2)$. In contrast, by applying the proposed sparse orthogonal gradient projection strategy, the memory complexity is reduced to $O((sC)^2)$, achieving a memory saving of $(1/s)^2$ times. This approach not only preserves the effectiveness of the original OGP in mitigating catastrophic forgetting but also significantly reduces the memory overhead, making it more suitable for incremental learning on resource-constrained edge devices.

\subsection{Overall Objective Function}
To sum up, the proposed method seeks to optimize the following minimization problem:
\begin{equation}
\mathcal{L} = \mathcal{L}_{ce} + \lambda \mathcal{L}_{proto},
\label{total_loss}
\end{equation}
where $\lambda$ is the trade-off hyper-parameter. $\mathcal{L}_{ce}$ is the classification cross entropy loss, which is formulated as:
\begin{equation}
\mathcal{L}_{ce} = -\sum_{(x, y) \in (X_t, Y_t)} y \log \mathcal{F}_g(\mathcal{F}_e(x)),
\end{equation}
where $(X_t, Y_t)$ is the dataset for task $\mathcal{T}_t$, $\mathcal{F}_e(\cdot)$ and $\mathcal{F}_g(\cdot)$ represents the feature extractor and classifier, respectively.
$\mathcal{L}_{proto}$ is the prototype loss in \cite{zhu2022self} to alleviate classifier catastrophic forgetting, which is expressed as:
\begin{equation}
\mathcal{L}_{proto} = -\sum_{(x, y) \in (X_{proto}, Y_{proto})} y \log \mathcal{F}_g(x),
\end{equation}
where $(X_{proto}, Y_{proto})$ denotes the class prototypes of the seen data.

\begin{algorithm}[t!]
	\DontPrintSemicolon
	\SetAlgoNoLine
        \small
	\KwIn {Training datasets $\{ X_t, Y_t\}$ for task $\mathcal{T}_t \in \{\mathcal{T}_1, \mathcal{T}_2, ...\}$, pre-trained network $f(\cdot,\mathbf{W}_0)$.}
	\KwOut {The network $f(\cdot,\mathbf{W})$ after learning from task sequence.}
        \textcolor{code_Color}{\# Step 1: \emph{Center-sensitive Kernel Selection and Center-decoupled Mechanism}}\\
        Decouple trainable layer parameters $\mathbf{W}$ into central kernel parameters $\mathbf{W}_{\alpha}$ and the remaining backbone parameters $\mathbf{W}_{\theta}$. 
        \graycomment{\emph{// Decoupled parameters as described in Eq. \ref{eq:3} and Eq. \ref{eq:4}}} \\
        Make $\mathbf{W}_{\alpha}$ trainable and $\mathbf{W}_{\theta}$ frozen \\
        \textcolor{code_Color}{\# Step 2: Train $f(\cdot,\mathbf{W}_0)$ on sequential tasks}\\
	\For{task $\mathcal{T}_t \in \{\mathcal{T}_1, \mathcal{T}_2, ...\}$}{
		\textcolor{code_Color}{\# Step 2.1: \emph{Dynamic Channel Element Selection}}\\
            Compute channel sensitivity $S_{\mathbf{W}_{\alpha}[c]}$ of central kernel elements $\mathbf{W}_{\alpha}$ and select the top $s$-proportion channels with the higher sensitivity. 
            \graycomment{\emph{// Central kernel elements sensitivity calculation as illustrated in Eq. \ref{channel_score}}}\\
            Make the selected central kernel element channels trainable and others frozen.\\
            \textcolor{code_Color}{\# Step 2.2: Train $f(\cdot,\mathbf{W}_0)$ on task $\mathcal{T}_t$}\\
            \While{not converaged}{
                Sample a batch $\{x, y\}$ from $\{ X_t, Y_t\}$. \\
                Forward propagation: $f(x,\mathbf{W}) = f(x,\mathbf{W}_{\alpha}) + f(x,\mathbf{W}_{\theta})$. \\
                Compute gradient $g$ of $\mathbf{W}_{\alpha}$ by $\mathcal{L}_{ce}$ and $\mathcal{L}_{proto}$. 
                \graycomment{\emph{// Total loss function as illustrated in Eq. \ref{total_loss}}} \\
                Project gradient $g$ to $g'$ by gradient projection space. 
                \graycomment{\emph{// Sparse Orthogonal Gradient Projection as shown in Fig. \ref{Framework}}} \\
                Update $\mathbf{W}_{\alpha}$ by $g'$.
            }
            Compute data prototype. \\
            Compute gradient projection space.
	}
	\caption{\textbf{C}enter-\textbf{s}ensitive \textbf{K}ernel \textbf{O}ptimization}
        \label{AlgorithmPNE-CL}
\end{algorithm}

\section{Experiments}~\label{Experiments}
\subsection{Experimental Setup}
\subsubsection{Datasets} 
We evaluate the efficiency and effectiveness of the proposed method on two representative benchmarks, CIFAR-100 \cite{krizhevsky2009learning} and TinyImageNet \cite{le2015tiny}. CIFAR-100 contains 100 classes, each with 500 training images and 100 testing images. TinyImageNet consists of 200 classes, each comprising 500 training images and 50 testing images. Following the experimental setup of recent incremental learning work \cite{zhu2021prototype, zhu2022self}, we initially train the model on half of the classes in the dataset, then divide the remaining data into either 5, 10, or 20 tasks for incremental training.

\subsubsection{Evaluation Metrics}
The incremental learning performance of the model is evaluated based on the average accuracy across all tasks.  To assess the efficiency of the method, we measure memory footprint \cite{wang2022sparcl} (mainly including model parameters, gradients, and activations during the training) and quantify the computational cost of training using Floating Point Operations Per Second (FLOPs) \cite{wang2022sparcl}.

\subsubsection{Baseline models}
To demonstrate the superiority of the proposed method in balancing performance and efficiency, we conduct two key comparisons. On the one hand, we compare our method with the recent pure low-cost training method \cite{yang2023efficient} to highlight its potential and advantages in training on resource-constrained edge devices. Additionally, we explore the result of directly applying an incremental learning strategy to the low-cost training method, illustrating the limitations of such a straightforward combination. On the other hand, we compare our proposed method with several representative conventional incremental learning methods under the same incremental settings. These conventional incremental learning methods typically prioritize improving incremental learning performance but impose significant computational and memory overhead, making them not friendly for resource-constrained edge devices. Furthermore, we compare the incremental learning performance and training resource overhead of our method against an existing low-cost incremental learning method. Overall, our focus is on demonstrating that the proposed method effectively balances model performance and training efficiency, making it a more viable solution for incremental learning on resource-constrained edge devices.

\subsubsection{Experimental Details}
The parameters are optimized using the Adam optimizer \cite{kingma2014adam}, with a weight decay of 0.0005. The initial learning rate is set to 0.001, which is reduced to 0.1 of the original every 45 epochs. The batch size is set to 128. We empirically set the channel sparsity rate $s$ to 75\%. All models are implemented within the PyTorch framework. To ensure a fair comparison, we strictly follow the incremental settings in prior works \cite{zhu2021prototype, zhu2022self} and use ResNet-18 as the backbone model. Following \cite{zhu2022self}, all models are pre-trained within 100 epochs, using only cross-entropy loss, and then each task is incrementally learned for 60 epochs, using cross-entropy loss and prototype balance loss. To ensure an equitable baseline for incremental learning, we match the same pre-trained accuracy of all datasets as \cite{zhu2022self}. We follow the training strategy of \cite{yang2023efficient} to better evaluate the feasibility of our method for incremental learning on resource-constrained edge devices, which involves selecting the final 2 or 4 layers for training.

\begin{table*}
	\centering
	\setlength{\tabcolsep}{9pt}
	\renewcommand{\arraystretch}{1.2} 
	\caption{Comparison of the \textit{average accuracy(\%)$\uparrow$} with low-cost training methods on CIFAR-100 and TinyImageNet datasets, along with \textit{computational cost (GFLOPs$\downarrow$)} and \textit{memory consumption (Mem (MB)$\downarrow$)} evaluated on CIFAR-100. $\mathrm{T}$ represents the number of incremental tasks. The proposed method achieves excellent incremental learning performance while meeting the low-cost training requirements on resource-constrained edge devices. }
	\begin{threeparttable}
		\begin{tabular}{c||ccc||ccc||ccc}
			\Xhline{1.2pt}
			\multirow{2.2}{*}{\fontsize{10}{10}{Method}}
			 &\multicolumn{3}{c||}{\rule{0pt}{10pt}\fontsize{10}{10}{CIFAR-100}} & \multicolumn{3}{c||}{\fontsize{10}{10}{TinyImageNet}} &
			\multirow{2.2}{*}{\fontsize{10}{10}{\makecell{Trainable\\Params (MB)}}} & \multirow{2.2}{*}{\fontsize{10}{10}{GFLOPs}} &
			\multirow{2.2}{*}{\fontsize{10}{10}{Mem (MB)}}\\
			\hhline{~||---||---||~~~}
			
			&$\mathrm{T}=5$&$\mathrm{T}=10$&$\mathrm{T}=20$&$\mathrm{T}=5$&$T=10$&$\mathrm{T}=20$\\ 
			\Xhline{1.0pt}
			
			GF \cite{yang2023efficient}
			&$22.91$ &$13.06$ &$8.43$  &$19.20$ &$11.53$ &$6.86$ &$18.20$  &$142.75$ &\cellcolor{gray!15} $76.34$\\ 
			\Xhline{1.0pt}
			
			GF \cite{yang2023efficient}+OGP \cite{wang2021training} 
			&$65.85$ &$65.03$ &$61.50$  &$50.70$ &$49.38$ &$48.64$ &$18.20$  &$142.75$ &\cellcolor{gray!15} $727.47$\\ 
			\Xhline{1.0pt}
			
			Ours 
			&$\boldsymbol{66.17}$ &$\boldsymbol{65.23}$ &$\boldsymbol{61.58}$  &$\boldsymbol{50.70}$ &$49.27$ &$\boldsymbol{48.65}$ &$\boldsymbol{2.20}$  &$144.90$ &\cellcolor{gray!15} $\boldsymbol{65.97}$\\ 
			\Xhline{1.2pt}
		\end{tabular}
	\end{threeparttable}
	\label{On_Device_Table}
\end{table*}

\begin{table*}
	\centering
	\setlength{\tabcolsep}{10pt}
	\renewcommand{\arraystretch}{1.2} 
	\caption{Comparison of the \textit{average accuracy(\%)$\uparrow$} with conventional incremental learning methods on CIFAR-100 and TinyImageNet, along with \textit{computational cost (GFLOPs$\downarrow$)} and \textit{memory consumption (Mem (MB)$\downarrow$)} evaluated on CIFAR-100. The proposed method significantly reduces memory and computational overhead with comparable performance to existing methods. }
	\begin{threeparttable}
		\begin{tabular}{c||ccc||ccc||ccc}
			\Xhline{1.2pt}
			\multirow{2.2}{*}{\fontsize{10}{10}{Method}}
			&\multicolumn{3}{c||}{\rule{0pt}{10pt}\fontsize{10}{10}{CIFAR-100}} & \multicolumn{3}{c||}{\fontsize{10}{10}{TinyImageNet}} &
			\multirow{2.2}{*}{\fontsize{10}{10}{\makecell{Trainable\\Params (MB)}}} &
			\multirow{2.2}{*}{\fontsize{10}{10}{GFLOPs}} &
			\multirow{2.2}{*}{\fontsize{10}{10}{Mem (MB)}}\\
			\hhline{~||---||---||~~~}
			
			& $\mathrm{T}=5$&$\mathrm{T}=10$&$\mathrm{T}=20$&$\mathrm{T}=5$&$\mathrm{T}=10$&$\mathrm{T}=20$\\ 
			\Xhline{1.0pt}
			
			\multicolumn{10}{c}{\textit{Conventional Incremental Learning}}\\
			\Xhline{1.0pt}
			
			LwF\_MC \cite{rebuffi2017icarl}  &$45.93$ &$27.43$ &$20.07$  &$29.12$ &$23.10$ &$17.43$ &$42.80$ &$426.11$ & \cellcolor{gray!15} $794.50$ \\ 
			\hhline{-||---||---||---}
			
			MUC \cite{liu2020more}  &$49.42$ &$30.19$ &$21.27$  &$32.58$ &$26.61$ &$21.95$ &$--$ &$--$ & \cellcolor{gray!15} $--$\\ 
			\hhline{-||---||---||---}
			
			PASS \cite{zhu2021prototype}  &$63.47$ &$61.84$ &$58.09$  &$49.55$ &$47.29$ &$42.07$ &$42.80$ &$568.32$ &\cellcolor{gray!15} $838.09$\\  
			\hhline{-||---||---||---}
   
               SSRE \cite{zhu2022self}  &$65.88$ &$65.04$ &$61.70$  &$50.39$ &$48.93$ &$48.17$ &$4.66$ &$614.91$ &\cellcolor{gray!15} $633.05$\\
               \hhline{-||---||---||---}
			
			SOPE \cite{zhu2023self}  &$66.64$ &$65.84$ &$61.83$  &$53.69$ &$52.88$ &$51.94$ &$--$  &$--$ &\cellcolor{gray!15} $--$\\ 
			\hhline{-||---||---||---}
			
			PR-AKA \cite{shi2023prototype}  &$70.02$ &$68.86$ &$65.86$  &$53.32$ &$52.61$ &$49.83$ &$42.80$ &$568.32$  &\cellcolor{gray!15} $838.09$\\ 
			\Xhline{1.0pt}
			
			\multicolumn{10}{c}{\textit{Low-cost Incremental Learning}}\\
			\Xhline{1.0pt}

            SparCL \cite{wang2022sparcl}  &$37.65$ &$25.83$ &$24.32$  &$23.20$ &$12.17$ &$5.67$ &$21.35$ &$426.11$  &\cellcolor{gray!15} $661.20$\\ 
            \hhline{-||---||---||---}
			Ours  &$66.17$ &$65.23$ &$61.58$  &$50.70$ &$49.27$ &$48.68$ &$\boldsymbol{2.20}$ &$\boldsymbol{144.90}$ &\cellcolor{gray!15} $\boldsymbol{65.97}$\\ 
			\Xhline{1.2pt}
			
		\end{tabular}
	\end{threeparttable}
	\label{Incremental_Table}
\end{table*}

\subsection{Quantitative Results and Comparisons}
\subsubsection{Comparison with Low-cost Training Methods}
Our experiments focus on validating the superiority of the proposed method in balancing efficiency and performance. As shown in Table \ref{On_Device_Table}, we conduct a detailed comparison of our method with the recent low-cost training method GF \cite{yang2023efficient}, demonstrating the potential of our method for training on resource-constrained edge devices. Low-cost training methods can facilitate efficient model training on resource-constrained edge devices. However, due to the lack of incremental learning capabilities in the pure low-cost training method, it can lead to catastrophic forgetting, resulting in typically poor performance in incremental learning. An intuitive idea is to apply incremental learning strategies to low-cost training methods. 

In this regard, we choose an Orthogonal Gradient Projection-based (OGP) incremental learning strategy \cite{wang2021training}, which is directly applied to GF and improves incremental learning performance. However, such conventional incremental learning strategies are usually resource-consuming like the SVD operation in the OGP above. In other words, the enhancement of incremental learning performance is accompanied by an increased demand for memory resources. This indicates the impracticality of the straightforward combination of incremental learning and low-cost training methods. In contrast, our method, while maintaining a resource expenditure comparable to GF, still significantly boosts the performance of incremental learning, with average improvements of 49.53\% and 37.01\% on CIFAR-100 and TinyImageNet, respectively. Compared to GF+OGP, the naive combination of low-cost training method GF and incremental learning strategy OGP, the proposed method can achieve comparable or even better performance with only 9\% of its memory overhead.

\subsubsection{Comparison with Conventional and Low-cost Incremental Learning Methods}
In this section, we conduct a comparative analysis of the proposed method with existing incremental learning methods, including conventional incremental learning and low-cost incremental learning methods. As shown in Table \ref{Incremental_Table}, compared with the \textit{conventional incremental learning} methods, our proposed method achieves comparable performance while significantly reducing resource (i.e. memory and computation) utilization. Specifically, it achieves a minimum of 3 times and 9 times compression in computational and memory resources, respectively. 

In Table 2, we show the comparison between our method and the existing \textit{low-cost incremental learning} method SparCL \cite{wang2022sparcl}, demonstrating the superiority of our method for low-cost incremental learning. Although SparCL attempts to reduce the resource burden by sparsifying parameters to reduce trainable parameters, its gradient computation still requires the participation of all kernel parameters, meaning that the reduction in trainable parameters does not fully translate into actual memory resource savings. Compared to SparCL, we only utilize 9.98\% and 34.01\% of its memory and computational overhead, outperforming it by an average of 35.06\% and 35.87\% on two benchmark datasets, respectively. 

\begin{table*}
	\centering
	\setlength{\tabcolsep}{4.2pt}
	\renewcommand{\arraystretch}{1.2} 
	\caption{Ablation study of the proposed method on TinyImageNet dataset. The two main components contributing to memory usage, Gradient Memory and OGP Memory, are reported separately.}
	\begin{threeparttable}
		\begin{tabular}{c|cc|ccc||ccc||ccc}
			\Xhline{1.2pt}
			\multicolumn{1}{c||}{\multirow{1.8}{*}{\rule{0pt}{10pt}{Configuration}}}
			& \multicolumn{2}{c||}{\rule{0pt}{10pt}\fontsize{10}{10}{Base Settings}}
			& \multicolumn{3}{c||}{\rule{0pt}{10pt}\fontsize{10}{10}{Components}}
			& \multicolumn{3}{c||}{\fontsize{10}{10}{TinyImageNet}}
			& \multirow{2.2}{*}{\fontsize{10}{10}{\makecell{Gradient\\Mem (MB)}}}
			& \multirow{2.2}{*}{\fontsize{10}{10}{\makecell{OGP\\Mem (MB)}}}
			& \multirow{2.2}{*}{\fontsize{10}{10}{\makecell{Total\\Mem (MB)}}}\\
			\hhline{~||--||---||---||~~~}
			
			\multicolumn{1}{c||}{} & \multicolumn{1}{c}{GF} & \multicolumn{1}{c||}{OGP} & CSKS & CDM & DCES & $\mathrm{T}=5$ & $\mathrm{T}=10$ & $\mathrm{T}=20$\\ 
			\Xhline{1.0pt}

            \multicolumn{1}{c||}{\circled{1}} & \multicolumn{1}{c}{\ding{51}} & \multicolumn{1}{c||}{} & & &   & $19.20$ & $11.53$ & $6.86$ &\cellcolor{gray!15} $18.20$ &\cellcolor{gray!15} $0.00$ & \cellcolor{gray!15} $76.34$\\
			
			\hhline{~||--||---||---||---}	
			
			\multicolumn{1}{c||}{\circled{2}} & \multicolumn{1}{c}{\ding{51}} & \multicolumn{1}{c||}{\ding{51}} & & &   & $50.70$ & $49.38$ & $48.64$ &\cellcolor{gray!15} $18.20$ (\textcolor{wine}{\(\uparrow \boldsymbol{0.00}\)}) &\cellcolor{gray!15} $648.00$ (\textcolor{wine}{\(\uparrow \boldsymbol{648.00}\)}) & \cellcolor{gray!15} $727.47$ (\textcolor{wine}{\(\uparrow \boldsymbol{651.13}\)})\\
			
			\hhline{~||--||---||---||---}	
			
			\multicolumn{1}{c||}{\circled{3}} & \multicolumn{1}{c}{\ding{51}} & \multicolumn{1}{c||}{} & \ding{51} & &   & $50.73$ & $49.27$ & $48.64$ &\cellcolor{gray!15}$18.20$ (\textcolor{dark_green}{\(\downarrow \boldsymbol{0.00}\)}) &\cellcolor{gray!15}$8.00$ (\textcolor{dark_green}{\(\downarrow \boldsymbol{640.00}\)})&\cellcolor{gray!15}$87.47$ (\textcolor{dark_green}{\(\downarrow \boldsymbol{640.00}\)})\\ 
			\hhline{~||--||---||---||---}
			
			\multicolumn{1}{c||}{\circled{4}} & \multicolumn{1}{c}{\ding{51}} & \multicolumn{1}{c||}{} & \ding{51} & \ding{51} &  & $50.73$ & $49.27$ & $48.64$  &\cellcolor{gray!15}$2.20$ (\textcolor{dark_green}{\(\downarrow \boldsymbol{16.00}\)})  &\cellcolor{gray!15}$8.00$ (\textcolor{dark_green}{\(\downarrow \boldsymbol{640.00}\)})& \cellcolor{gray!15}$73.47$ (\textcolor{dark_green}{\(\downarrow \boldsymbol{654.00}\)})\\ 
			\hhline{~||--||---||---||---}
			\multicolumn{1}{c||}{\circled{5}} & \multicolumn{1}{c}{\ding{51}} & \multicolumn{1}{c||}{} & \ding{51} & \ding{51} & \ding{51}   & $50.70$ & $49.27$ & $48.65$ &\cellcolor{gray!15}$\boldsymbol{2.20}$ (\textcolor{dark_green}{\(\downarrow \boldsymbol{16.00}\)}) &\cellcolor{gray!15}$\boldsymbol{0.50}$ (\textcolor{dark_green}{\(\downarrow \boldsymbol{647.50}\)}) &\cellcolor{gray!15} $\boldsymbol{65.97}$ (\textcolor{dark_green}{\(\downarrow \boldsymbol{661.50}\)}) \\				
			\Xhline{1.2pt}
			
		\end{tabular}
	\end{threeparttable}
	\label{Ablation_Table}
\end{table*}

\subsection{Ablation Study}
\subsubsection{Effectiveness of Key Components}
As shown in Fig. \ref{Framework}, the proposed CsKO framework mainly contains the following three components: Center-sensitive Kernel Selection (CsKS) and Center-decoupled Mechanism (CdM), Dynamic Channel Element Selection (DCES). Here, we conduct comprehensive ablation studies on them to justify their superiority in memory saving. From the experimental results shown in Table \ref{Ablation_Table}, we can make the following observations.

\textit{Configuration \circled{1}: Low-cost training method without any incremental learning strategies.} Low-cost training method GF proposes activation and gradient block strategy to reduce the training memory, which can achieve efficient training. However, when dealing with continuously arriving new data, this method exhibits limitations in learning new information while keeping learned knowledge. Therefore, this lack of incremental learning capability leads to poor performance in low-cost incremental settings.

\textit{Configuration \circled{2}: Low-cost training method with naive application of incremental learning strategy.}  To achieve efficient and effective low-cost incremental learning, a naive implementation is to apply an incremental learning strategy to a low-cost training method. In this configuration, we directly integrate the OPG incremental learning strategy into the low-cost training method GF. As shown in Table \ref{Ablation_Table}, the introduction of OPG significantly enhances the performance of incremental learning. However, the components introduced by existing incremental learning methods to obtain superior performance are frequently resource-consuming, such as the OPG strategy above with a memory footprint of 648MB. Overall, the memory consumption of this configuration is 9.5 times higher than that of the low-cost training method in Configuration \circled{1}, making it unfriendly to training on resource-constrained edge devices.

\textit{Configuration \circled{3}: The proposed low-cost incremental learning method with only CSKS.} The introduction of the CSKS mechanism into GF demonstrates performance closely comparable to Configuration \circled{2}, which indicates that the model performance can be maintained with only a small number of central kernel elements. This supports the finding of our empirical analysis that the central kernel elements exhibit a stronger ability for new knowledge learning. More importantly, we develop sparse OGP based on the CSKS mechanism, which ensures incremental learning performance while reducing memory overhead to 1.2\% of the conventional OGP.

\textit{Configuration \circled{4}: The proposed low-cost incremental learning method with CsKS and CdM.} CdM collaborates with CsKS to provide a low-cost optimization mechanism called CsKO, which can further reduce memory costs while maintaining model performance. Specifically, CdM decouples the central kernel elements from the backbone and forms a side branch, allowing only the gradient flow branch to perform independent gradient calculation. Overall, this configuration achieves a 90\% savings in memory consumption under the CsKO mechanism.

\textit{Configuration \circled{5}: The proposed low-cost incremental learning method with CsKS, CDdM and DECS.} We further introduce the proposed online dynamic channel element selection strategy into the method, which involves learning through some selected channels that are more important to incoming data. It further promotes sparse OGP, reducing the final memory consumption of OGP to less than 0.1\% of the original.

Based on the above ablation experiments, we can conclude that our method can achieve the dual benefits of performance and efficiency by optimizing a small number of carefully selected parameters. These results align with existing research \cite{ding2023parameter} that underscores the pivotal role of efficient parameter subsets for superior performance.

\begin{figure}[htbp]
	\centering	
        \includegraphics[width=\linewidth]{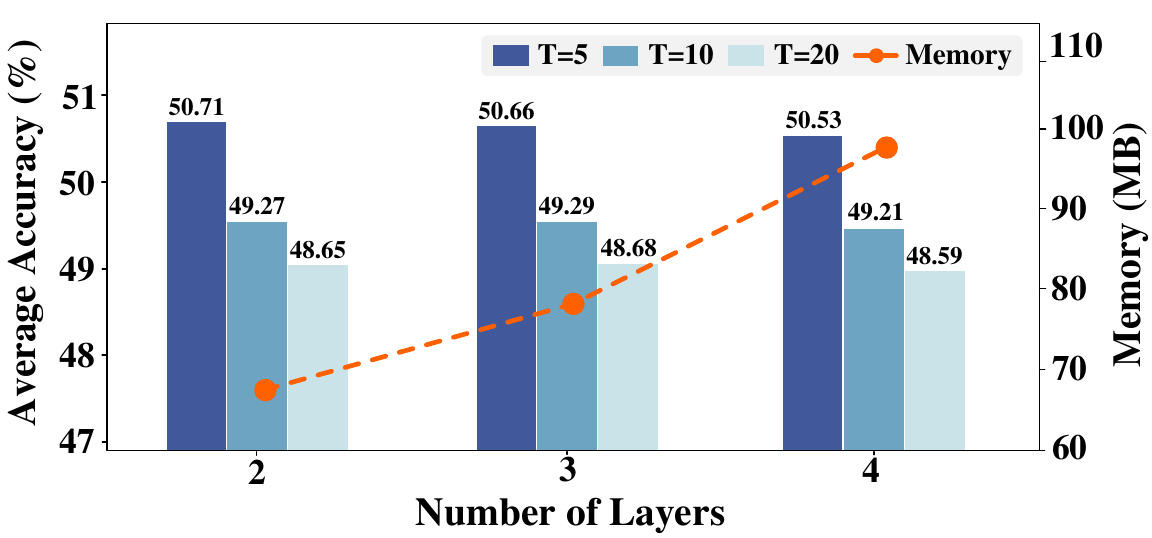}
	\caption{Performance under the different number of last trainable layers on TinyImageNet dataset.}
	\label{Layer_Num}
\end{figure}

\subsubsection{Analysis on Different Number of Trainable Layers}
We explore the impact of different numbers of last trainable layers on model performance and memory cost, as shown in Fig. \ref{Layer_Num}. The dashed line represents the average performance of three incremental task settings with different numbers of trainable network layers. It shows that there is no significant change in performance as the number of trainable layers increases. The solid line indicates the memory overhead for different numbers of trainable layers, which exhibits an increasing trend as the number of trainable layers grows. This implies that adding more trainable layers raises training costs without further performance gains. This may be because more parameters involved in learning new tasks also lead to the forgetting of a lot of historical knowledge. Therefore, we choose the central kernel elements of the last two layers to learn new knowledge to achieve the optimal balance between model performance and training cost.

\begin{figure}[htbp]
	\centering
	\begin{subfigure}{\linewidth}
		\includegraphics[width=\linewidth]{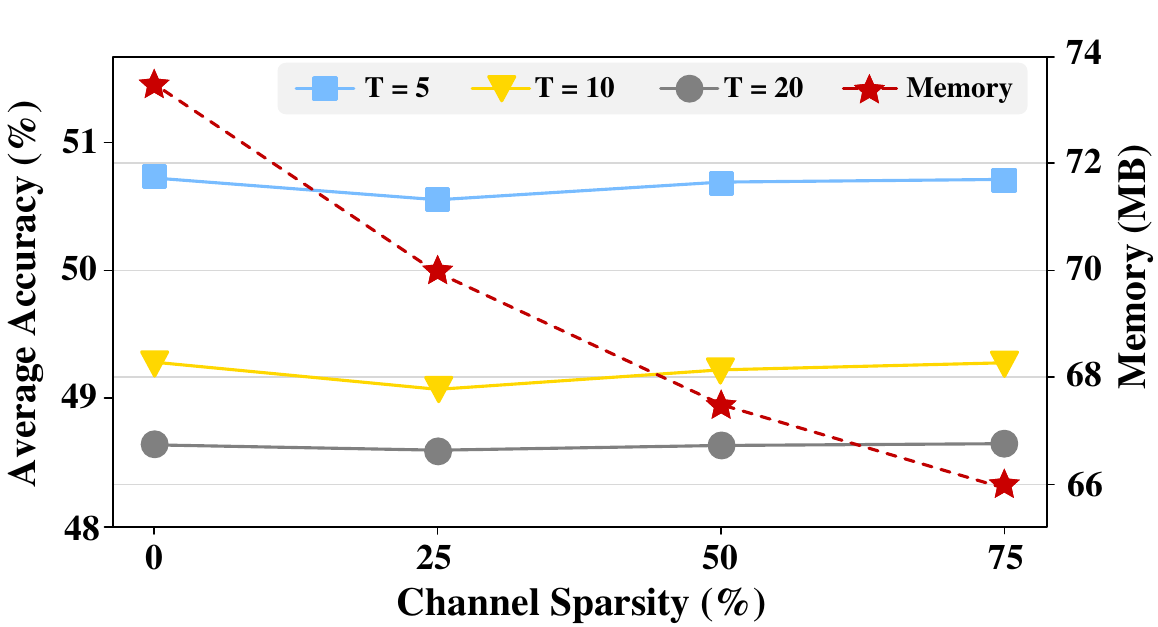}
	\end{subfigure}
	\caption{Performance under different channel sparsity in dynamic channel selection strategy on TinyImageNet dataset. Channel sparsity represents the proportion of sparsed (frozen) channels in total channels.}
	\label{Channel_Sparsity}
\end{figure}

\subsubsection{Analysis on Different Channel Sparsity}
We explore the effects of different channel sparsity on model performance and memory consumption, with channel sparsity defined as the ratio of sparse channels to the total number of channels. As shown in Fig. \ref{Channel_Sparsity}, under different channel sparsity, the performance fluctuates slightly compared to the full-channel training (i.e., 0\% channel sparsity setting), indicating that the performance-efficiency balance can be further promoted by appropriate setting of channel sparsity. Here, we set the channel sparsity rate at 75\%, with performance comparable to full-channel training performance, while achieving a 10\% memory savings.

\begin{table}[htbp!]
	\centering
        \setlength{\tabcolsep}{11pt}
	\renewcommand{\arraystretch}{1.2}
	\caption{Performance under different criteria for dynamic channel element selection on TinyImageNet dataset. Layer $i$ represents the $i$-th last convolution layer of the model. "New/Old" refers to the selection according to the sensitivity of channels towards new/old task data.}
	\begin{threeparttable}
		\begin{tabular}{cc||ccc}
			\Xhline{1.2pt}
			\multicolumn{2}{c||}{\rule{0pt}{10pt}\fontsize{10}{10}{Criteria}}
			 & \multicolumn{3}{c}{\fontsize{10}{10}{TinyImageNet}}  \\
			\hhline{--||---}
			
			Layer $1$ &Layer $2$  &$\mathrm{T}=5$ &$\mathrm{T}=10$ &$\mathrm{T}=20$\\ 
			\Xhline{1.0pt}
			
			Old &Old   &$50.60$ &$49.31$ &$48.70$  
			\\
			\hhline{--||---}
			
			New &Old  &$50.64$ &$49.27$ &$48.68$  
			\\ \hhline{--||---}
			
			Old &New  &$50.63$ &$49.27$ &$48.64$   \\
			\hhline{--||---} 	
			
			New &New &$50.70$ &$49.27$ &$48.65$  \\  	
			\Xhline{1.2pt}
			
		\end{tabular}
	\end{threeparttable}
	\label{CD_Table}
\end{table}

\subsubsection{Exploration on Channel Elements Selection Criteria}
We explore the impact of choosing different channel selection criteria for different network layers. Here, we focus on the last two layers of the network as a case study, and the results are shown in Table \ref{CD_Table}, "New/Old" refers to the selection according to the sensitivity of channels towards new/old task data. If channels that are more sensitive to new task data are selected, these channels are employed for the acquisition of new knowledge, while the remaining channels are frozen. This selection paradigm will lean more towards the plasticity of the model. On the other hand, if channels exhibit a significant sensitivity to old task data, these channels are rendered inactive to conserve the old knowledge deemed to be of greater importance, while the remaining channels are used for learning new knowledge. This selection strategy favors the stability of the model. We can find that the choice of different criteria for channel selection does not result in significant performance differences. This can be attributed to the inherent differences between new and old data, which theoretically leads to the selection of mutually exclusive channels. This implies that we always allocate channels to new and old tasks in a rational manner, thereby ensuring that they have access to channels that are deemed most crucial to their respective tasks.

\subsubsection{Exploration on Different Kernel Shapes}
\begin{figure}
	\centering
	\begin{subfigure}{\linewidth}
		\includegraphics[width=\linewidth]{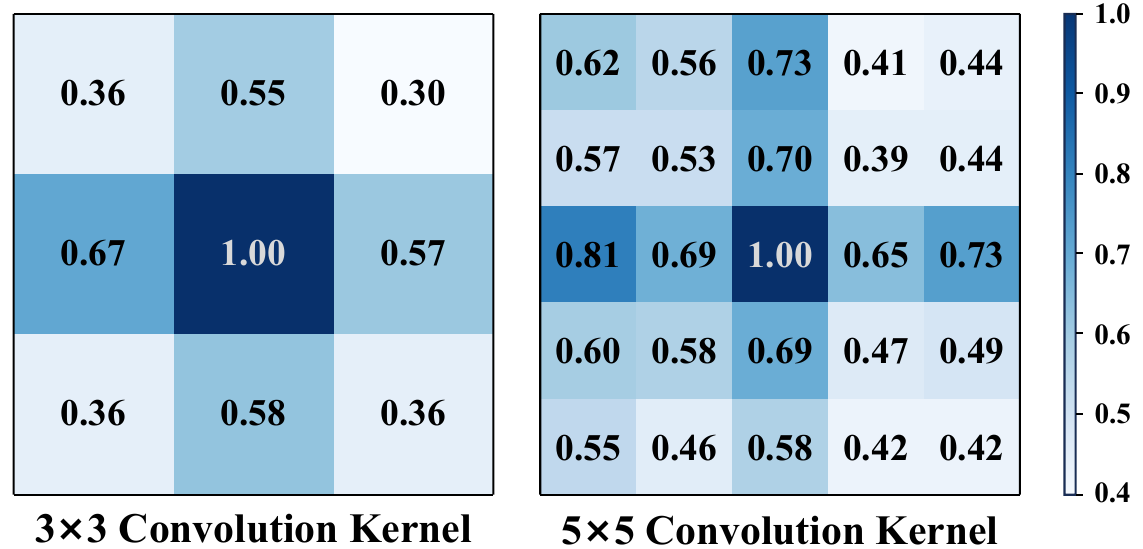}
	\end{subfigure}
	\caption{Sensitivity analysis results for $3 \times 3$ and $5 \times 5$ convolutional kernel in AlexNet. The values in the figure represent the sensitivity scores corresponding to the spatial positions of the convolutional kernels.}
	\label{Alex}
\end{figure}

To further validate our findings, we extend our exploration to other shapes of convolution kernels, such as the $5 \times 5$ convolution kernel utilized in AlexNet \cite{KrizhevskySH12}. 
Fig. \ref{Alex} shows the spatial sensitivity results of $3 \times 3$ convolution and $5 \times 5$ convolution in AlexNet. We can find that in different networks and different sizes of convolution kernels, the central position always shows more importance, which further supports our findings. 

\begin{figure}[htbp]
	\centering
	\begin{subfigure}{\linewidth}
		\includegraphics[width=\linewidth]{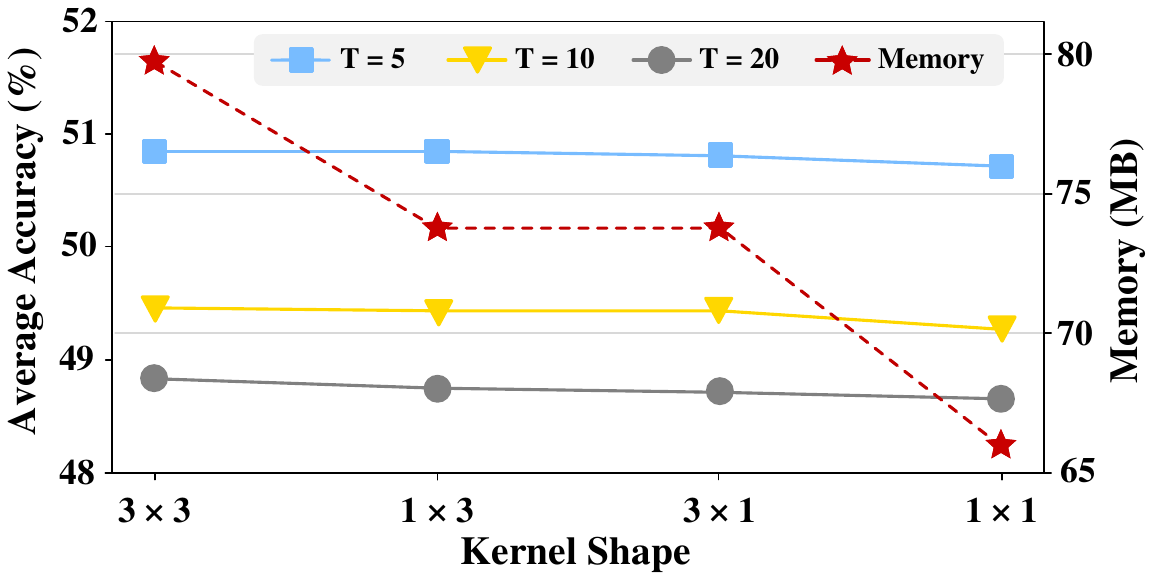}
	\end{subfigure}
	\caption{Performance under different kernel shapes as trainable parameters on TinyImageNet dataset.}
	\label{Kernel_Shape}
\end{figure}

At the same time, we also have a new observation, that is, the cross position of the convolution kernel tends to show more significant importance than the corner position. 
This observation prompts us to question whether selecting elements from the cross positions for learning new knowledge could potentially enhance performance.
To answer this question, we conduct an experiment with four distinct element selection strategies: $3 \times 3$ convolution, $1 \times 3$ convolution, $3 \times 1$ convolution, and $1 \times 1$ convolution. As shown in Fig. \ref{Kernel_Shape}, compared to the $1 \times 1$ convolution, the performance gain of $3 \times 3$ convolution, $1 \times 3$ convolution and $3 \times 1$ convolution is limited, with a 21\% additional memory overhead yields only a 0.17\% improvement in performance. This underscores the advantage of the proposed central kernel element optimization mechanism in achieving a balanced trade-off between performance and training efficiency.

\begin{figure}
	\centering	
        \includegraphics[width=\linewidth]{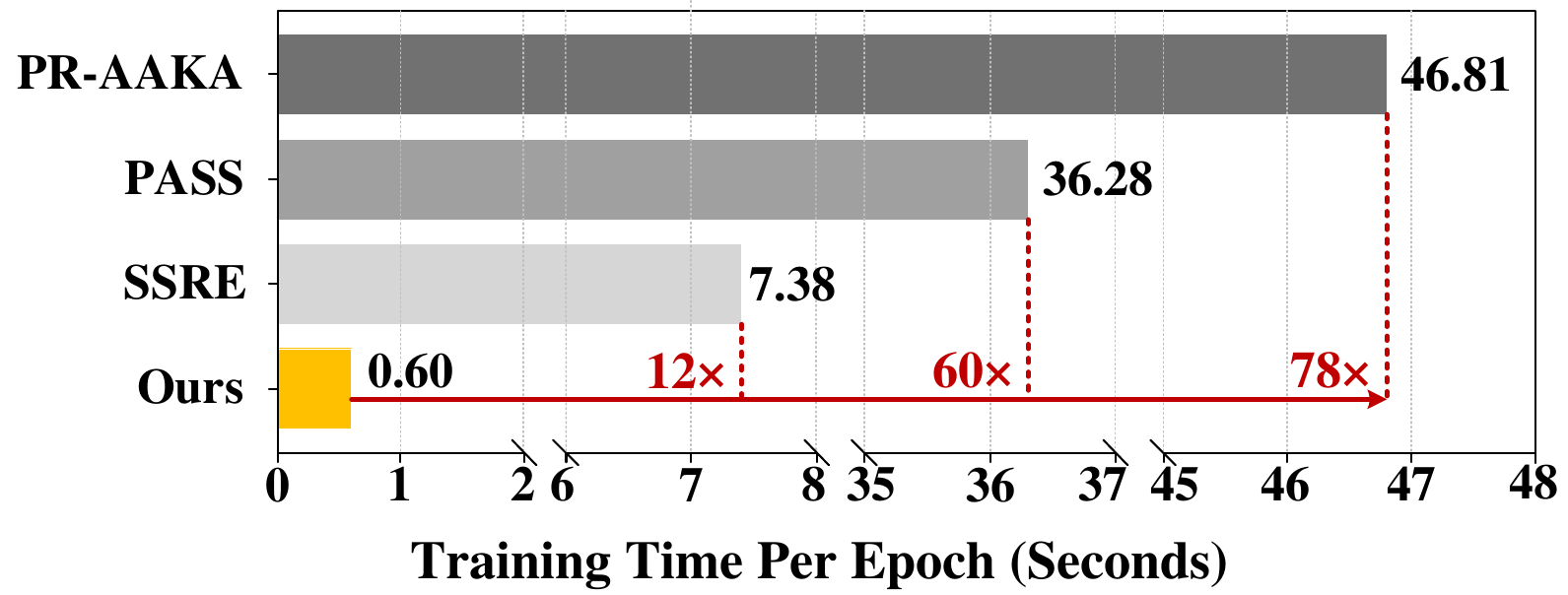}
	\caption{Comparison of training time with existing incremental learning methods on 32 $\times$ 32 sized images in CIFAR-100 dataset. Training time per epoch is calculated by averaging the training time duration over 10 epochs with a batch size of 128.}
	\label{Training_Time}
\end{figure}

\subsection{Training Cost Analysis}
\subsubsection{Memory Analysis}
The proposed method considers the primary factors contributing to the significant memory overhead during training and introduces targeted strategies to address these challenges. Specifically, under the proposed learning mechanism supported by the center-sensitive kernel optimization and dynamic channel element selection strategies, trainable parameters can achieve independent gradient computation at the parameter level. Therefore, the gradient of the trainable parameters $\tilde{\mathbf{W}}_{\alpha}^{i} \in \mathbb{R}^{D \times (sC) \times 1 \times 1}$ for the $i$-th layer is quantitatively limited to $D_{i} \times (sC_{i})$, thereby achieving a memory saving of around $\frac{K \times K}{s}$ times in terms of gradient, where $s \in [0, 1]$ denotes sparsity.

GF \cite{yang2023efficient} is a method to facilitate efficient training on edge devices, which employs a patch approximation strategy to reduce memory usage of activation. It simplifies the input feature $\mathbf{X}^{i} \in \mathbb{R}^{B \times C \times H \times W}$ to approximated $\mathbf{X}_{\textit{a}}^{i} \in \mathbb{R}^{B \times C \times \lceil \frac{H}{r} \rceil \times \lceil \frac{W}{r} \rceil}$, where $r$ the size of patch.
Despite its effectiveness in facilitating cost-efficient training, it encounters the challenge of catastrophic forgetting when learning new knowledge incrementally. Drawing inspiration from this, we adopt GF as our foundational method and propose an efficient and effective low-cost incremental learning framework. Given that the proposed method is orthogonal to GF, it inherently benefits from the advantage of reducing the memory overhead of activations.

In this study, the implementation of the incremental learning strategy based on orthogonal gradient projection necessitates additional memory allocation. This arises due to the intrinsic characteristics of conventional orthogonal gradient projection-based incremental learning strategies, which impose a substantial memory overhead primarily resulting from the Singular Value Decomposition (SVD) operation applied to the covariance matrix. However, benefiting from the proposed center-sensitive kernel optimization mechanism coupled with the dynamic channel element selection strategy, we further develop a sparse orthogonal gradient projection incremental learning strategy. Notably, this strategy preserves the efficacy of the original version while incurring only a minimal memory overhead. 
Theoretically, in the conventional orthogonal gradient projection-based method, the memory complexity of the SVD operation of the covariance matrix $\mathbf{M} \in \mathbb{R}^{C \times C}$ is $O({C}^2)$. In the proposed method, the memory complexity of sparse SVD is only $O((sC)^2)$, achieving a memory saving of $(1/s)^2$ times.

\subsubsection{Training Time Analysis}
We evaluate the proposed method against existing incremental learning methods in terms of training time overhead, as shown in Fig. \ref{Training_Time}. These experiments are conducted on 128 images of 32 $\times$ 32, with the average training time over 10 epochs used to represent the training time per epoch. For existing incremental learning methods, the training time per epoch typically ranges from several tens of seconds, with PR-AAKA requiring 46.8 seconds and PASS taking 36.3 seconds. The SSRE method achieves reduced training time overhead by freezing the backbone network and updating only the small side branches, resulting in a lower training time of 7.4 seconds per epoch. In contrast, our method requires merely 8\% of the training time overhead of SSRE, completing one epoch in just 0.6 seconds. The lower training time reflects the lower computational cost of the proposed method, which can enable resource-constrained edge devices to save more power consumption.

\begin{figure}[htbp]
	\centering	
        \includegraphics[width=0.95\linewidth]{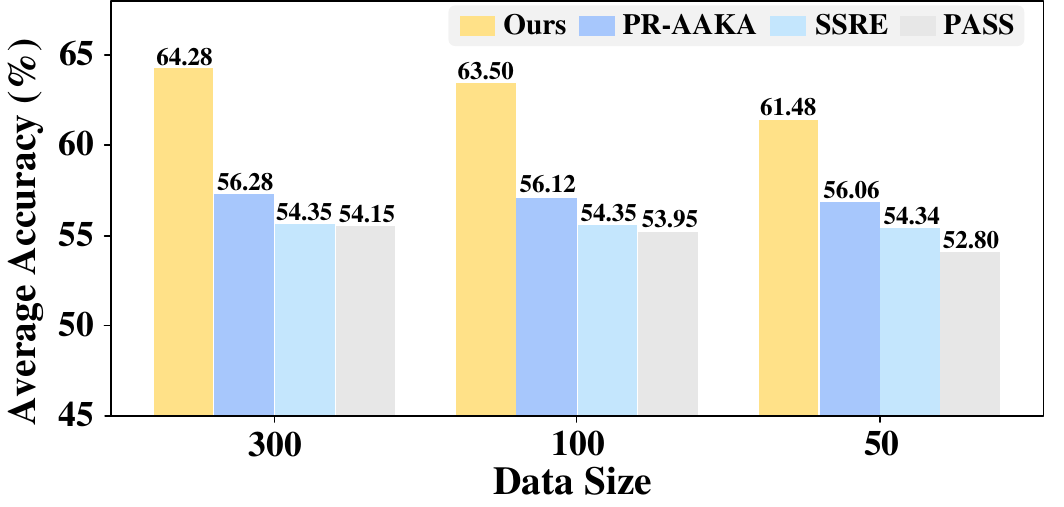}
	\caption{Comparison of the average accuracy(\%)$\uparrow$ with existing incremental learning methods under low-shot experiment setting on CIFAR-100 dataset. ``Data Size'' refers to the amount of data per class in the incremental sessions. }
	\label{Low_Shot}
\end{figure}

\subsection{Low-Shot Incremental Learning}
Typically, models are pre-trained on the server using large datasets and then deployed to edge devices. However, new classes frequently emerge in this dynamically changing world, requiring the model to continuously learn new knowledge while retaining old knowledge to adapt to new environments. Edge devices often operate in various complex real-world scenarios, one common scenario being the scarcity of newly acquired training samples. For example, facial recognition systems may only collect a few images of new users due to privacy concerns. The insufficiency of new class training data can lead to overfitting on the small amount of new data, preventing the model from adequately learning new knowledge and causing significant catastrophic forgetting of previously learned information.

To evaluate the effectiveness of the proposed method in low-shot scenarios on edge devices, we conduct experiments under the low-shot incremental learning setting and compare the proposed method with existing incremental learning methods in terms of average accuracy, computational cost and total memory usage. In the low-shot incremental learning setup, the pre-training session is provided with sufficient training data for each class (e.g., thousands or hundreds of images per class), while the incremental sessions are characterized by extremely limited data (typically hundreds or dozens of images per class). As shown in Fig. \ref{Low_Shot}, the proposed method always shows better performance than other incremental learning methods in low-shot scenarios.

\section{Conclusion}~\label{Conclusion}
In conclusion, this study proposes a simple yet effective low-cost incremental learning framework. Our empirical study finds that the central kernel is pivotal for maximizing knowledge intensity when learning new data, while freezing other kernels can effectively balance new knowledge learning and catastrophic forgetting. We further propose a center-sensitive kernel optimization framework and dynamic channel element selection strategies to significantly reduce the cost of gradient calculations and back-propagation. Besides, the proposed dynamic channel element selection strategy facilitates a sparse orthogonal gradient projection, further reducing the optimization burden. Extensive experiments demonstrate our method is efficient and effective, indicating its potential for advancing edge intelligence in dynamic environments. Future work will continue to explore the performance and applicability of low-cost incremental learning.

\bibliographystyle{IEEEtran}
\bibliography{Main}

\end{document}